%% file: kalofolias_2014.tex
\documentclass{article}
\usepackage{nips14submit_e, times}  
\usepackage{url}
\usepackage{subcaption}

\usepackage{amsmath}
\usepackage{amsfonts}
\usepackage{xcolor}

\usepackage{todonotes}

\input{vassilis_macros}

\title{Matrix Completion on Graphs}

\author{
Vassilis Kalofolias\\
Ecole Polytechnique F\'ed\'erale de Lausanne\\
\texttt{vassilis.kalofolias@epfl.ch}
\And
Xavier Bresson\\
Ecole Polytechnique F\'ed\'erale de Lausanne\\
\texttt{xavier.bresson@epfl.ch}
\And
Michael Bronstein\\
Universit\`a della Svizzera Italiana\\
\texttt{michael.bronstein@usi.ch}
\And
Pierre Vandergheynst\\
Ecole Polytechnique F\'ed\'erale de Lausanne\\
\texttt{pierre.vandergheynst@epfl.ch}
}

\nipsfinalcopy

\begin{document}

\maketitle

\begin{abstract}
\vspace{-0.15cm}
The problem of finding the missing values of a matrix given a few of its entries, called matrix completion, has gathered a lot of attention in the recent years. Although the problem under the standard low rank assumption is NP-hard, Cand\`es and Recht showed that it can be exactly relaxed if the number of observed entries is sufficiently large. In this work, we introduce a novel matrix completion model that makes use of proximity information about rows and columns by assuming they form communities. This assumption makes sense in several real-world problems like in recommender systems, where there are communities of people sharing preferences, while products form clusters that receive similar ratings. Our main goal is thus to find a low-rank solution that is structured by the proximities of rows and columns encoded by graphs. We borrow ideas from manifold learning to constrain our solution to be smooth on these graphs, in order to implicitly force row and column proximities. Our matrix recovery model is formulated as a convex non-smooth optimization problem, for which a well-posed iterative scheme is provided. We study and evaluate the proposed matrix completion on synthetic and real data, showing that the proposed structured low-rank recovery model outperforms  the standard matrix completion model in many situations.
\end{abstract}

\vspace{-0.1cm}
\section{Introduction}
\vspace{-0.2cm}
How to reconstruct signals exactly from very few measurements? This central question in signal processing has been extensively studied in the last few years and has triggered a fast emerging field of research, namely compressed sensing. Exact recovery from few measurements is actually possible if the signal is {\it sparse} in some representation domain. A related essential question has been recently considered for matrices: is it possible to reconstruct matrices exactly from very few observations? It appears that exact recovery is also possible in this setting if the matrix is {\it low-rank}. The problem of low-rank recovery from sparse observations is referred as the {\it matrix completion} problem. Several important real-world problems can be cast as a matrix completion problem, including remote sensing \cite{art:Schmidt86RemoteSensing}, system identification \cite{art:LiuVandenberghe09SysIden} and recommendation systems \cite{srebro2004maximum}. Throughout the paper, we will consider the recommendation system problem as an illustration of the matrix completion problem.
Such systems have indeed become very common in many applications such as movie or product recommendation (e.g. Netflix, Amazon, Facebook, and Apple). 
The Netflix recommendation system tries to predict ratings of movies never seen by users. Collaborative filtering is widely used today to solve this problem \cite{breese1998empirical}, inferring recommendations by finding similar rating patterns and using them to complete missing values. This is typically a matrix completion problem where the unknown values of the matrix are computed by finding a low-rank matrix that fits the given entries.

How much information is needed for the exact recovery of low-rank matrices? 
In the case of random uniformly sampled entries without noise, Cand{\`e}s and Recht showed in \cite{candes2009exact} that, to guarantee perfect recovery, the number of observed entries must be larger than $c n^{1.2}r\log n$ for $n\times n$ matrices of rank $r$ (this bound has been refined more recently, see \cite{recht2011simpler} and references therein). 
The case of noisy observations was studied in \cite{candes2010matrix,negahban2012restricted}, while a non-uniform sampling setting was considered in \cite{salakhutdinov2010collaborative}. 
In this work, we propose to use additional information about rows and columns of the matrix to further improve the matrix completion solution.

In the standard matrix completion problem, rows and columns are assumed to be completely {\it unorganized}. However, in many real-world problems like the Netflix problem, there exist relationships between users (such as their age, gender, hobbies, education, etc) and movies (such as their genre, release year, actors, origin country, etc). This information can be taken advantage of, since people sharing the same tastes for a class of movies are likely to rate them similarly. We make use of {\it graphs} to encode relationships between users and movies and we introduce a new reconstruction model called {\it matrix completion on graphs}. Our main goal is to find a low-rank matrix that is {\it structured} by the proximities between users and movies. 
%
Introducing structure in sparse recovery problems is not new in the literature of compressed sensing  \cite{pro:HuangZhangMetaxas09StrucSpars,art:BaraniukCevherDuarteHegde10SparStru, art:JenattonAudibertBach11SparStru}, while similar structure inducing regularization has been proposed for factorized models for matrix completion \cite{ma2011recommender}. Yet, introducing structures via graphs in the convex low-rank matrix recovery setting is novel to the best of our knowledge. 
We note that a large class of recommendation systems, called content-based filtering, use graphs and clustering techniques to make predictions \cite{huang2002graph}. Along this line, our proposed methodology can be seen as a hybrid recommendation system that combines collaborative filtering (low-rank property) and content-based filtering (graphs of users and movies).  

%
We borrow ideas from the field of manifold learning \cite{belkin2001laplacian,belkin2003laplacian} and force the solution to be smooth on the manifolds of users and movies. We make the standard assumption that the graphs of users and movies are (non-uniform) discretizations of the corresponding manifolds. Forcing smoothness can be achieved through different regularizations. We use the popular Dirichlet/Laplacian energy, whose minimizing flow is the well-known linear heat diffusion on manifold/graph, and show that the proposed model leads to a convex {\it non-smooth} optimization problem. Convexity is a desired property for uniqueness of the solution. Non-smoothness can be highly challenging, however, our problem belongs to the class of $\ell_1$-type optimization problems, for which several recently proposed efficient solvers exist \cite{boyd2011distributed,combettes2011proximal,nesterov2013first}.  The corresponding algorithm is derived in Section \ref{secOpt}. It is tested on synthetic and real data \cite{pro:Miller03MovieLens} in Section \ref{secNumExp}.

\vspace{-0.1cm}
\section{Original matrix completion problem}
\label{secMatCom}
\vspace{-0.2cm}
The problem of matrix completion is to find the values of an $m\times n$ matrix $M$ given a sparse set $\Omega$ of observations $M_{ij} : (i,j) \in \Omega \subseteq \{1,\hdots, m\} \times \{1, \hdots, n\}$. 
Problems of this kind are often encountered in collaborative filtering or recommender system applications, the most famous of which is the Netflix problem, in which one tries to predict the rating that $n$ users (columns of $M$) would give to $m$ films (rows of $M$), given only a few ratings provided by each user. 
A particularly popular model is to assume that the ratings are affected by a few factors, resulting in a low-rank matrix. 
This leads to the rank minimization problem 
\vspace{-0.1cm} 
\begin{eqnarray}
\min_{X \in \mathbb{R}^{m\times n}} 			 \text{rank}(X)   \hspace{3mm} \text{s.t.} \hspace{3mm} {\cal A}_{\Omega}(X) = {\cal A}_{\Omega}(M), 
\label{eq:matcomp_np}
\end{eqnarray}
where ${\cal A}_{\Omega}(M) = (M_{ij \in \Omega})$ denotes the observed elements of $M$. Problem~(\ref{eq:matcomp_np}) is NP-hard. However, replacing $\text{rank}(X)$ with its convex surrogate known as the {\em nuclear} or {\em trace norm} \cite{srebro2004maximum}  $\| X\|_* = \mathrm{tr}((XX^\top)^{1/2}) = \sum_{k} \sigma_k$, where $\sigma_k$ are singular values of $X$, one obtains a semidefinite program 
\vspace{-0.1cm}
\begin{eqnarray}
\min_{X \in \mathbb{R}^{m\times n}} 			 \|X\|_*    \hspace{3mm} \text{s.t.} \hspace{3mm} {\cal A}_{\Omega}(X) = {\cal A}_{\Omega}(M).  
\label{eq:matcomp_nuc}
\end{eqnarray}
Under the assumption that $M$ is sufficiently incoherent, if the indices $\Omega$ are uniformly distributed and $|\Omega|$ is sufficiently large, the minimizer of~(\ref{eq:matcomp_nuc}) is unique and coincides with the minimizer of~(\ref{eq:matcomp_np}) \cite{candes2009exact,recht2011simpler}. If in addition the observations are contaminated by noise, one can reformulate problem~(\ref{eq:matcomp_nuc}) as 
\vspace{-0.3cm}
\begin{eqnarray}
\min_{X \in \mathbb{R}^{m\times n}}	\gamma_n \|X\|_*  +  \ell\left({\cal A}_{\Omega} (X),{\cal A}_{\Omega}(M)\right)  ,
\label{eq:matcomp_nuc1}
\end{eqnarray}
where the data term $\ell$ in general depends on the type of noise assumed. If $\ell$ is the squared Frobenius norm $\|A_\Omega\circ (X-M)\|_F^2$ ($A_\Omega$ here is the observations mask matrix, $\circ$ the Hadamard product), the distance between the solution of ~(\ref{eq:matcomp_nuc1}) and $M$ can be bounded by the norm of the noise \cite{candes2010matrix}. 

One notable disadvantage of problems~(\ref{eq:matcomp_nuc}-\ref{eq:matcomp_nuc1}) is the assumption of a ``good'' distribution of the observed elements $\Omega$, which implies, in the movie rating example, that on average each user rates an equal number of movies, and each movie is rated by an equal number of users. In practice, this uniformity assumption is far from being realistic: for instance, in the Netflix dataset, the number of movie ratings of different users varies from $5$ to $10^4$. When the sampling is non-uniform, the quality of the lower bound on $|\Omega|$ deteriorates dramatically \cite{salakhutdinov2010collaborative}, from approximately constant number of observations per row in the former case, to an order of $n^{1/3} - n^{1/2}$ in the latter. In such settings, Salakhutdinov and Srebro \cite{salakhutdinov2010collaborative} suggest using the weighted nuclear norm $\| X \|_{*(p,q)} = \| \mathrm{diag}(\sqrt{p}) X \mathrm{diag}(\sqrt{q}) \|_*$, where $p$ and $q$ are $m$- and $n$- dimensional row- and column-marginals of the distribution of observations, showing a significant performance improvement over the unweighted nuclear norm. Pathologically non-uniform sampling patterns, such as an entire row or column of $M$ missing, cannot be handled. Furthermore, in many situations the number of observations might be significantly smaller than the lower bounds. 

%

\vspace{-0.1cm}
\section{Matrix completion on graphs}
\label{secMatComOnGrap}
\vspace{-0.2cm}

Low rank implies the linear dependence of rows/columns of $M$. However, this dependence is unstructured. In many situations, the rows/columns of matrix $M$ possess additional structure that can be incorporated into the completion problem in the form of a regularization. In this paper, we assume that rows/columns of $M$ are given on vertices of graphs. In the Netflix example, the users (columns of $M$) are the vertices of a ``social graph'' whose edges represent e.g. friendship or similar tastes  relations. Thus, it is reasonable to assume that connected users would give similar movie ratings, i.e., interpreting the ratings as an $m$-dimensional vector-valued function on the $n$ vertices of the social graph, such a function would be {\em smooth}.

More formally, let us be given the undirected weighted {\em row graph} $G_r = (V_r, E_r, W_r)$ with vertices $V_r = \{1, \hdots, m\}$ and edges $E_r \subseteq V_r \times V_r$ weighted with non-negative weights represented by the $m \times m$  matrix $W_r$; and respectively the {\em column graph} $G_c = (V_c, E_c, W_c)$ defined in the same way.  Let $X \in \mathbb{R}^{m\times n}$ be a matrix, which we will regard as a collection of $m$-dimensional column vectors denoted with subscripts $X = (x_1, \hdots, x_n)$, or of $n$-dimensional row vectors denoted with superscripts $X = ((x^1)^\top, \hdots,  (x^m)^\top)^\top$. Regarding the columns $x_1, \hdots, x_n$ as a vector-valued function defined on the vertices $V_c$, the smoothness assumption implies that $x_j \approx x_{j'}$ if $(j,j') \in E_c$. Stated differently, we want 
\vspace{-0.15cm}
\begin{eqnarray}
\sum_{j,j'} w^c_{jj'} \| x_j - x_{j'} \|_2^2 \quad = \quad \tr(X L_c X^\top) \quad = \quad \|X\|^2_{\mathcal{D},c}
\label{eq:priorc}
\end{eqnarray}
\vspace{-0.07cm}to be small, where $D_c=\mathrm{Diag}(\sum_{j'=1}^n w^c_{jj'})$, $L_c = D_c - W_c$ is the Laplacian of the column graph $G_c$, and $\|\cdot\|_{\mathcal{D},c}$ is the graph Dirichlet semi-norm for columns. Similarly, for the rows we get a corresponding expression $\tr(X^\top L_r X)=\|X\|^2_{\mathcal{D},r}$ with the Laplacian $L_r$ of the row graph $G_r$. These smoothness terms are added to the matrix completion problem as regularization terms (in the sequel, we treat the case where $\ell$ is the squared Frobenius norm), 
\vspace{-0.07cm}
\begin{align}
&\min_X 				\gamma_n\|X\|_* + \ell\left({\cal A}_{\Omega}(X), {\cal A}_{\Omega}(M)\right) +  \frac{\gamma_r}{2} \|X\|^2_{\mathcal{D},r} + \frac{\gamma_c}{2} \|X\|^2_{\mathcal{D},c}. \label{eq:problem2}
\end{align}
\vspace{-0.55cm}
\paragraph{Relation to simultaneous sparsity models.}
Low rank promoted by the nuclear norm implies sparsity in the space of outer products of the singular vectors, i.e., in the singular value decomposition $X  =  \sum_{k} \sigma_k u_k v_k^\top$ only a few coefficients $\sigma_i$ are non-zero. Recent works \cite{oymak2012simultaneously} proposed imposing additional structure constraints, considering matrices that are simultaneously low-rank (i.e., sparse in the space of singular vectors outer products) and sparse (in the original representation). Our regularization can also be considered as a kind of simultaneously structured model. The column smoothness prior~(\ref{eq:priorc}) makes the rows of $X$ be close to the eigenvectors of the column graph Laplacian $L_c$, i.e., each row of $X$ can be expressed as a linear combination of a few eigenvectors of $L_c$. This can be interpreted as row-wise sparsity of $X$ in the column graph Laplacian eigenbasis. 
Similarly, the row smoothness prior results in column-wise sparsity of $X$ in the row graph Laplacian eigenbasis. Overall, the whole model~(\ref{eq:problem2}) promotes simultaneous sparsity of $X$ in the singular vectors outer product space, and row/column-wise sparsity in the respective Laplacian eigenspaces.


\vspace{-0.05cm}
\section{Optimization}
\vspace{-0.15cm}
\label{secOpt}
{\bf Algorithm.} Problems like \eqref{eq:problem2} containing non-differential terms cannot be tackled efficiently with a direct approach, while proximal based methods can be applied. We use the \textit{Alternating Direction Method of Multipliers (ADMM)} that has seen great success recently \cite{boyd2011distributed} (other choices could include f.e. \textit{fast iterative soft thresholding} \cite{beck2009fast}) by first introducing the equivalent splitting version of \eqref{eq:problem2}
\vspace{-0.1cm}
\begin{eqnarray}
\min_{X,Y\in\mathbb{R}^{m\times n}}\  \underbrace{ \gamma_n \|X\|_* }_{F(X)} + \underbrace{ \frac{1}{2}\|A_\Omega \circ (Y-M)\|_\mathcal{F}^2 + \frac{\gamma_r}{2} \|Y\|^2_{\mathcal{D},r} + \frac{\gamma_c}{2} \|Y\|^2_{\mathcal{D},c} }_{G(Y)} \quad \textrm{s.t.}\quad X=Y. 
\label{eqSplit}
\end{eqnarray}
This splitting step followed by an augmented Lagrangian method to handle the linear equality constraint is what constitutes ADMM. The success of ADMM for $\ell_1$ problems is mainly due to the fact that it does not require an exact solution for the iterative sub-optimization problems, but rather an {\it approximate} solution. The augmented Lagrangian of \eqref{eqSplit} is $\mathcal{L}(X,Y,Z) = F(X) + G(Y) + \langle Z,X-Y\rangle + \frac{\rho}{2} \|X-Y\|_\mathcal{F}^2$. Both $F$ and $G$ are closed, proper and convex, and since we have no inequality constraints Slater's conditions and therefore strong duality hold. Then $(X^\star,Y^\star)$ and $Z^\star$ are primal-dual optimal if $(X^\star,Y^\star,Z^\star)$ is a saddle point of the augmented Lagrangian $\mathcal{L}$, i.e. $\sup_{Z} \inf_{X,Y} \mathcal{L}(X,Y,Z) = \mathcal{L}(X^\star,Y^\star,Z^\star) = \inf_{X,Y} \sup_{Z}  \mathcal{L}(X,Y,Z),$ or $\mathcal{L}(X^\star,Y^\star,Z) \leq \mathcal{L}(X^\star,Y^\star,Z^\star) \leq \mathcal{L}(X,Y,Z^\star)\quad \forall X,Y,Z$. ADMM finds a saddle point with the following iterative scheme
\vspace{-0.2cm}
\begin{eqnarray}
X^{k+1} &=& \argmin_X \mathcal{L}(X,Y^k,Z^k) \label{eqADMMx}, 
\end{eqnarray}
\vspace{-0.5cm}
\begin{eqnarray}
Y^{k+1} &=& \argmin_Y \mathcal{L}(X^{k+1},Y,Z^k) \label{eqADMMy}, 
\end{eqnarray}
\vspace{-0.5cm}
\begin{eqnarray}
Z^{k+1} &=& Z^k + \rho ( X^{k+1} - Y^{k+1} ) \label{eqADMMz}.
\vspace{-0.1cm}
\end{eqnarray}
Eventually the convergence of the proposed ADMM algorithm \eqref{eqADMMx}-\eqref{eqADMMz} can be studied (and likely proved) with different mathematical approaches, for example \cite{cai2010singular}.


{\bf Solving sub-optimization problems.} ADMM algorithms can be very fast as long as we can compute fast approximate solutions to the sub-optimization problems, here \eqref{eqADMMx} and \eqref{eqADMMy}. \\
Problem \eqref{eqADMMx} requires finding $X^{k+1}$ that minimizes $\mathcal{L}(X,Y^k,Z^k)$, i.e. $X^{k+1} = \argmin_X \gamma_n \|X\|_* + \rho/2 \|X-H\|_\mathcal{F}^2,\quad \textrm{where}\quad H=Y^k-\rho^{-1}Z^k=\textrm{prox}_{ F/\rho } (H)$, where $\textrm{prox}_E$ is the proximal operator of $E$ defined as $\textrm{prox}_E(H)=\argmin_X E(X) + 1/2 \|X-H\|_\mathcal{F}^2$. In the case of the nuclear norm, there exists a closed-form solution: $X^{k+1} = U \textrm{soft}_{\gamma_n/\rho}(\Lambda) V^\top$, where $U,V,\Lambda$ are respectively the singular value decomposition (SVD) of $H$, i.e. $H=U \Lambda V^\top$, and $\textrm{soft}_\eta(\lambda)=\max(0,\lambda-\eta)\frac{\lambda}{|\lambda|}$ is the soft-thresholding operator defined for $\lambda\in\mathbb{R}$ \cite{cai2010singular}.\\
Problem \eqref{eqADMMy} requires finding $Y^{k+1}$ that minimizes $\mathcal{L}(X^{k+1},Y,Z^k)$, i.e. $Y^{k+1} = \argmin_Y 1/2\|A_\Omega \circ (Y-M)\|_\mathcal{F}^2 + \gamma_r/2 \|Y\|^2_{\mathcal{D},r} + \gamma_c/2 \|Y\|^2_{\mathcal{D},c} + \rho/2 \|Y-H\|_\mathcal{F}^2 = \textrm{prox}_{ G/\rho } (H),\quad \textrm{where}\quad H=X^{k+1}+\rho^{-1}Z^k$. Unlike Problem \eqref{eqADMMx}, there is no closed-form solution to compute the proximal of $G$ as the solution is given by solving a linear system of equations. Precisely, the optimality condition of \eqref{eqADMMy} is $A_\Omega \circ (Y-M) + \gamma_r Y L_r + \gamma_c L_c Y + \rho (Y - H) = 0$, which can be re-written as $Ay=b$ as follows: $( \tilde{A}_\Omega + \gamma_r   L_r \otimes I_n + \gamma_c  I_m \otimes L_c + \rho I_{mn} ) \textrm{vec}(Y) = \textrm{vec} ( M + \rho H )$, where $\textrm{vec}(.)$ is the column-stack vectorization operator, $\otimes$ is the Kronecker product, $\tilde{A}_\Omega=\textrm{Diag}(\textrm{vec}(A_\Omega))$ and we use the formula $\textrm{vec}(ABC) = (C^\top \otimes A) \textrm{vec}(B)$. Also note that $A$ is symmetric positive semidefinite (s.p.s.d) as the Kronecker product of two s.p.s.d matrices, thus the conjugate gradient (CG) algorithm can be applied to compute a fast approximate solution of \eqref{eqADMMy}.


{\bf Computational complexity.} The overall complexity of the algorithm is dominated by the computation of the nuclear proximal solutions by SVD, whose complexity is $O(mn^2)$
per iteration for $m>n$ \cite{golub2012matrix}. The computational complexity of the CG algorithm is $O(kmn)$
for $k$-NN graphs.

\vspace{-0.05cm}
\section{Numerical experiments}
\vspace{-0.15cm}

\subsection{Synthetic `Netflix' dataset}
\vspace{-0.25cm}
\begin{figure}
\label{fig:artificial_data}
          \begin{subfigure}[b]{.5\linewidth}
            \centering\includegraphics[scale=.28]{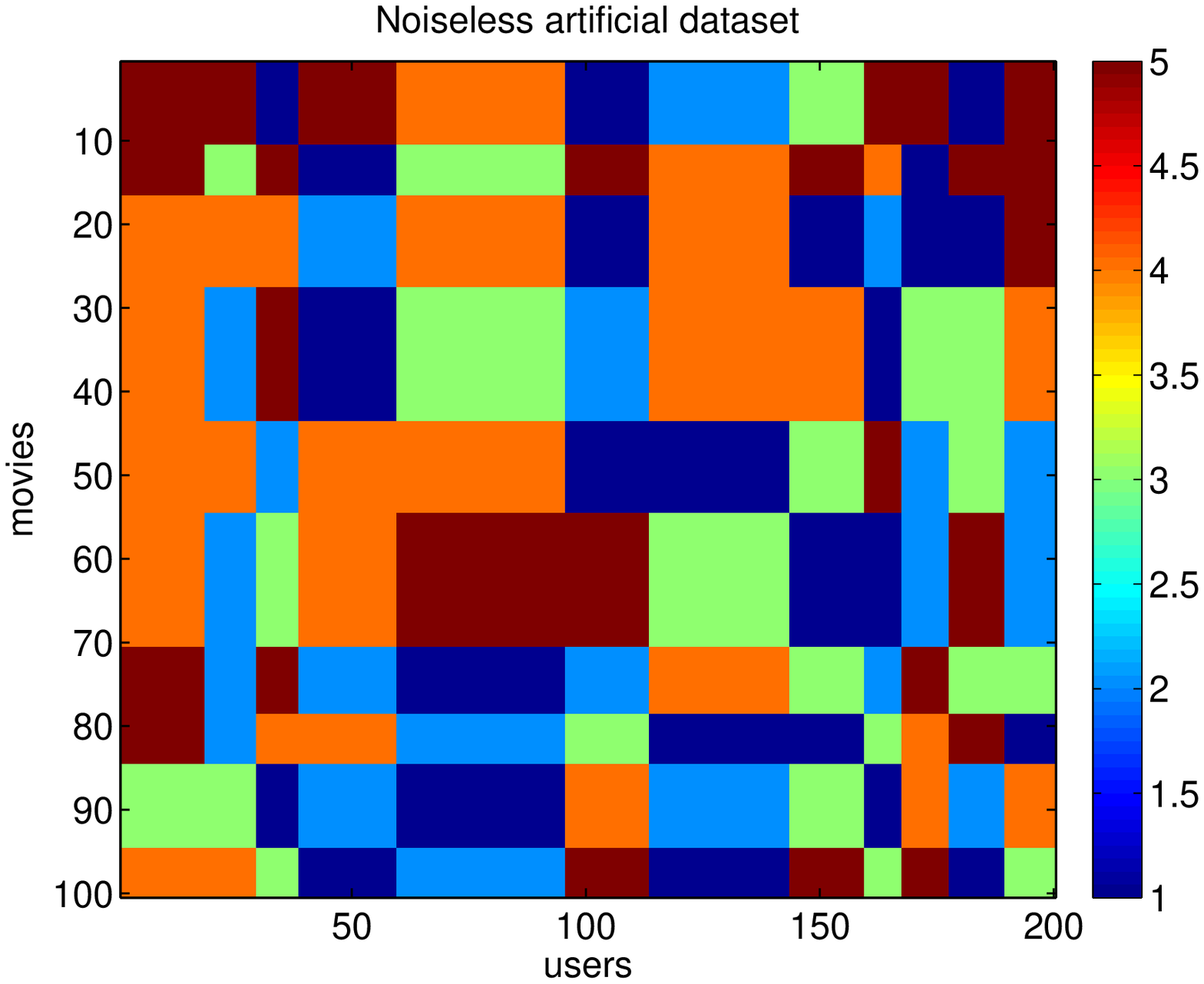}
            \vspace{-0.25cm}
            \caption{Rank-10 matrix $M$ of ideal ratings}\label{fig:art_M_noiseless}
          \end{subfigure}
           \begin{subfigure}[b]{.5\linewidth}
            \centering \includegraphics[scale=.28, trim=10 10 40 10, clip=true]{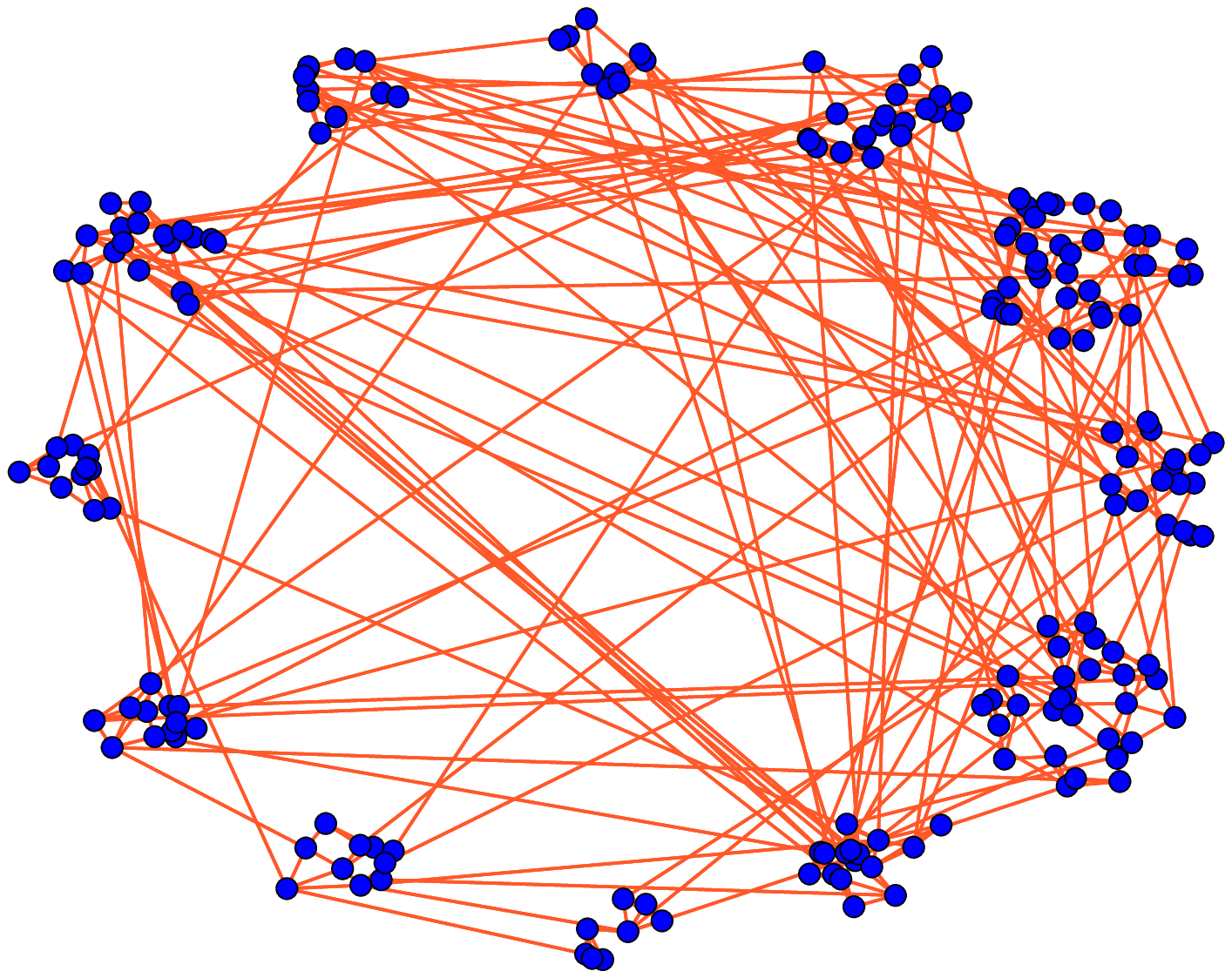}
            \vspace{-0.25cm}
            \caption{Graph of users with $25\%$ of wrong edges}\label{fig:artificial_G_u}
          \end{subfigure}
        \caption{Synthetic `Netflix' dataset}
	\vspace{-0.25cm}
\end{figure}
We start the evaluation of our matrix recovery model with a synthetic Netflix-like dataset, to study the behavior of model under controlled conditions. 
The artificial dataset $M$ is generated such that if fulfills two assumptions: (1) $M$ is \textit{low-rank} and (2) its columns and rows are respectively \textit{smooth} w.r.t. the column graph $G_c$ and the row graph $G_r$. 
Figure \ref{fig:art_M_noiseless} shows our synthetic dataset. It is inspired by the problem of movie recommendations as elements of $M$ are chosen to be integers from $\{1\hdots 5\}$ like in the Netflix prize problem. The matrix in Fig. \ref{fig:art_M_noiseless} is \textit{noiseless}, showing the ideal ratings for each pair of user and movie groups.

The row graph $G_r$ of the matrix $M$ is constructed as follows. The rows of $M$ are grouped into $10$ \textit{communities} of different sizes. We connect nodes within a community using a $3$-nearest neighbors graph and then add different amounts of erroneous edges, that is, edges between vertices belonging to different communities. The erroneous edges form a standard Erd\H{o}s-R\'{e}nyi graph with variable probability. We follow the same construction process for the column graph $G_c$ that contains $12$ communities. For both graphs, binary edge weights are used. The intuition behind this choice of graphs is that \textit{users form communities of people with similar taste}. Likewise, movies can be grouped according to their type, so that \textit{movies of the same group obtain similar ratings}. The users graph is depicted in Fig \ref{fig:artificial_G_u}, where nodes of the same community are clustered together. Note that matrix $M$ in Fig. \ref{fig:art_M_noiseless} has rank equal to the minimum of user communities and the movie communities, in this case $10$. 

\vspace{-0.1cm}
\subsubsection{Recovery quality versus number of observations}
\begin{figure}
\label{fig:art_uniform_all}
          \begin{subfigure}[b]{.5\linewidth}
            \centering \includegraphics[scale=.38, trim=0 0 0 0, clip=true]{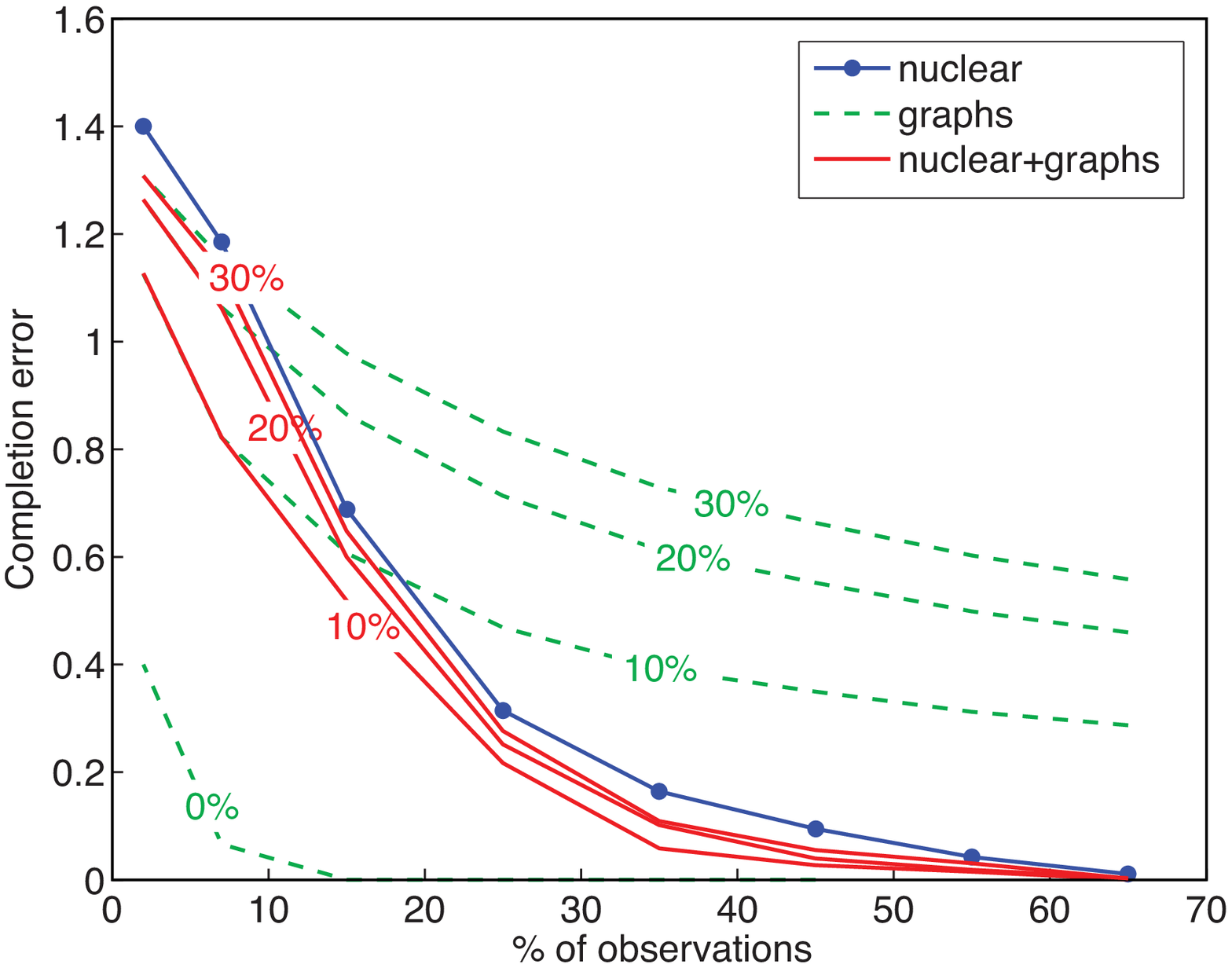}
            \caption{Noiseless observations}\label{fig:art_uniform_noiseless}
          \end{subfigure}
          \begin{subfigure}[b]{.5\linewidth}
            \centering\includegraphics[scale=.38, trim=0 0 0 0, clip=true]{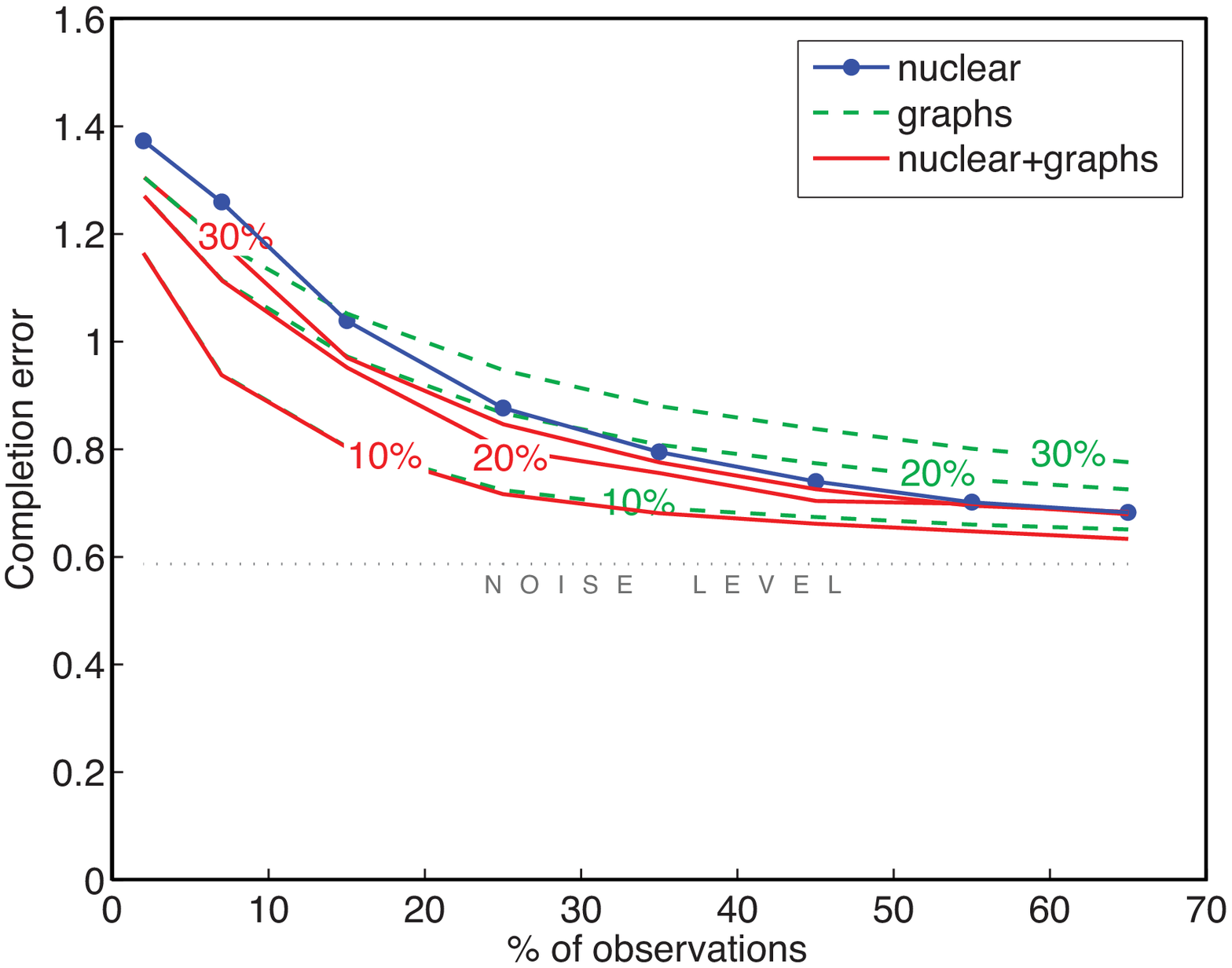}
            \caption{Noisy observations}\label{fig:art_uniform_noisy}
          \end{subfigure}
        \caption{Matrix recovery error on synthetic `Netflix' dataset (uniform sampling). Percentage of erroneous edges in graphs is shown on top of green and red lines. }
	\vspace{-0.45cm}
\end{figure}

\vspace{-0.25cm}
Two standard assumptions often made in the literature on matrix completion are that the observed elements of the matrix are sampled \textit{uniformly at random}, and that the reconstructed matrix is \textit{perfectly} low-rank (the case that we call \textit{noiseless}).
{\bf Noiseless case. } We test the performance of our method in this setting, comparing it to the standard nuclear norm-based matrix completion (a particular case of our problem with $\gamma_c = \gamma_r = 0$) and to a method that uses only the graphs ($\gamma_n = 0$). We reconstruct the matrix $M$ using different levels of observed values and report the reconstruction root mean squared error (RMSE) on a fixed set of $35\%$ of the elements that was not observed. The result is depicted in Fig \ref{fig:art_uniform_noiseless}. 
We use graphs with $10\%$, $20\%$, and $30\%$ of erroneous edges. 
Noisy graphs alone (green lines) perform poorly compared to the nuclear norm reconstruction (blue line). However, when we use both graphs and nuclear norm (red lines), we obtain results that are better than any of the two alone.\\
\begin{figure}
\label{fig:art_non_uniform_all}
          \begin{subfigure}[b]{.5\linewidth}
            \centering \includegraphics[scale=.38, trim=0 0 0 0, clip=true]{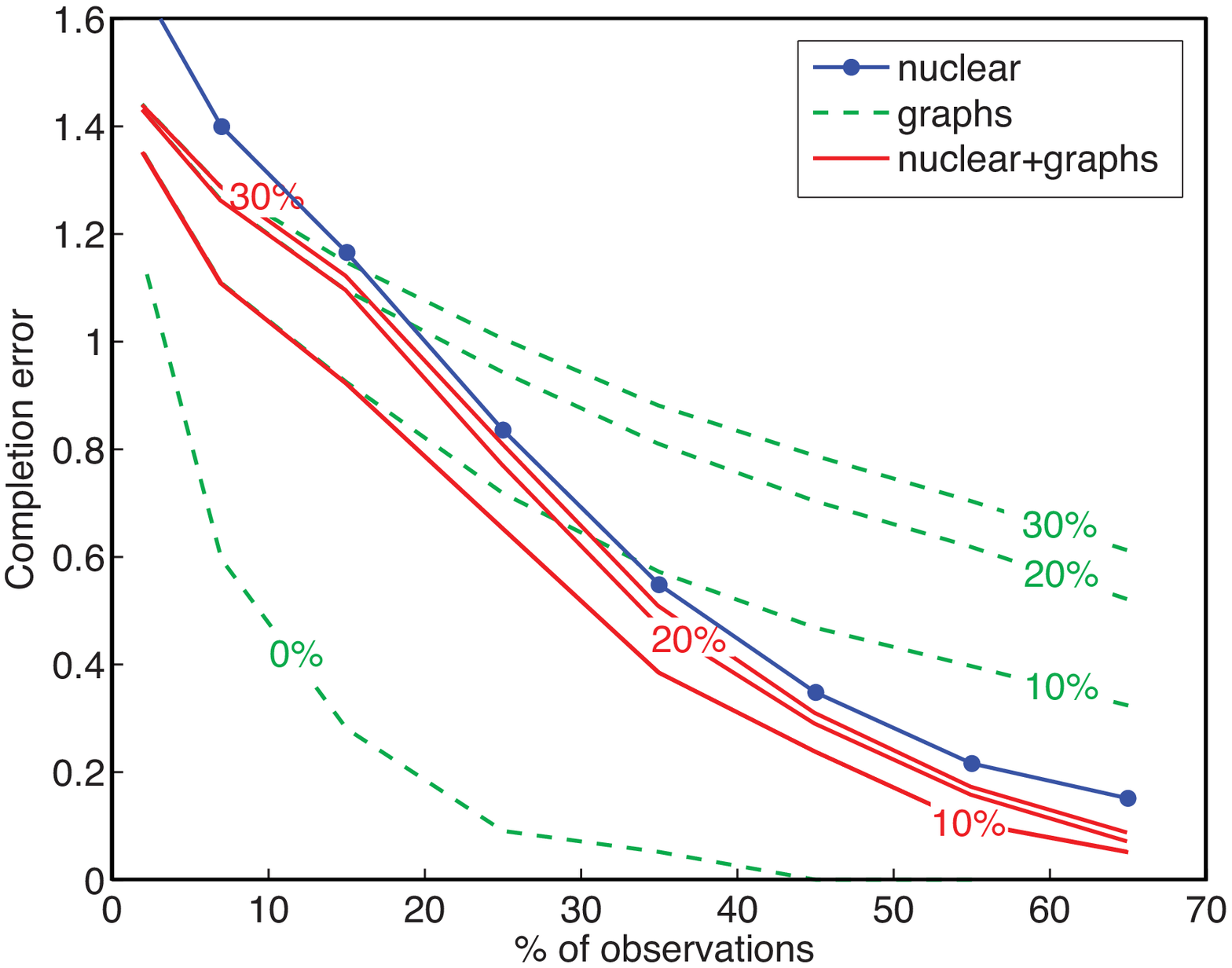}
            \caption{Noiseless observations}\label{fig:art_non_uniform}
          \end{subfigure}
          \begin{subfigure}[b]{.5\linewidth}
            \centering\includegraphics[scale=.38, trim=0 0 0 0, clip=true]{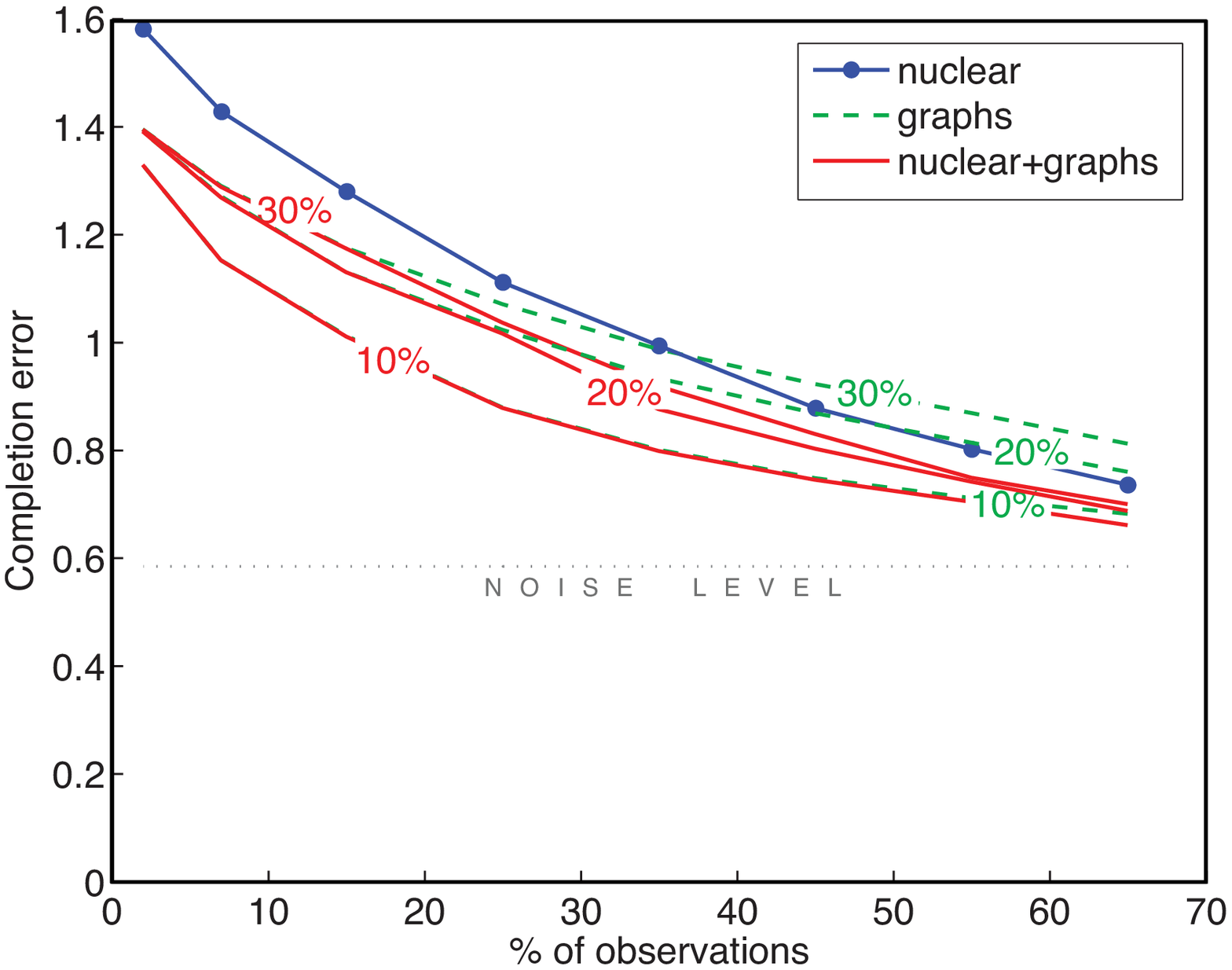}
            \caption{Noisy observations}\label{fig:art_non_uniform_noisy}
          \end{subfigure}

        \caption{Matrix recovery error on synthetic `Netflix' dataset (non-uniform sampling). Percentage of erroneous edges in graphs is shown on top of green and red lines.}
	\vspace{-0.45cm}
\end{figure}
{\bf Noisy case. } We add noise to $M$ using a discretized Laplacian distribution. This type of noise models the human tendency to impulsively over- or under-rate a movie. In this case, the matrix that we try to reconstruct is \textit{close} to low-rank, and the nuclear norm is still expected to perform well. As we see in Fig. \ref{fig:art_uniform_noisy} though, if we have high-quality graphs (green line with $10\%$ erroneous edges), we can expect the same reconstruction quality of the nuclear norm regularization by using just \textit{half of the number of observations} and only with the graph smoothness terms (green dashed line), that computationally is much cheaper to run. In this figure, the dashed black line designates the level of added noise in the data.\\
Note also that even if we use connectivity information of relatively bad quality (green dashed line with $30\%$ wrong edges), we can still benefit by combining the smoothness and the low-rank regularization terms (solid red line). Therefore the combination of nuclear and graph is robust to graph construction errors for low levels of observation. However, when the observation level is high enough ($>50\%$ for this \textit{specific} size of matrices - note that this number may vary significantly depending on the matrix size), this benefit is lost (solid red line) and the nuclear norm regularizer (blue line) works better without the graph smoothness terms.

{\bf Non-uniform sampling. } 
As noted in \cite{salakhutdinov2010collaborative}, the pattern of the observed values in real datasets does not usually follow a uniform distribution. In fact, the observations are such that the rating frequencies of users and movies closely follow \textit{power law distributions}. 
In our experiment, we assume a simple generative process where users and movies are independently sampled from a power law, that is $pr\left(\textrm{sample} \{i,j\}\right) = 1/ ij$. This is a very sparse distribution with fixed expected number of observations, so we repeat this process identically $s$ times in order to control the overall density. Our final sampling is the logical OR operator of all these $s$ `epochs', that follows the distribution
$p\left(\{i,j\} \in \Omega\right) = 1 - \left(1 - 1/ij\right)^s. $
We find that this simple sampling scheme gives results close to the actual ratings of real datasets such as the MovieLens 10M that we use in the following.\\
The results of our experiments for this setting are summarized in Fig. \ref{fig:art_non_uniform}. Not surprisingly, all methods suffer from the non-uniformity of the sampling distribution. Still, the nuclear norm (blue line) crosses the line of a high-quality graph ($10\%$ green line) only after $35\%$ observations, while in the uniform case, Fig. \ref{fig:art_uniform_noiseless}, this happened for less than $20\%$ observed values. A similar behavior is exhibited for the noisy case, Fig. \ref{fig:art_non_uniform_noisy}. There, the nuclear norm regularization quality is better than the medium-quality graph ($20\%$ green line) only for more than $45\%$ observations, while in the uniform case, Fig. \ref{fig:art_uniform_noisy}, the corresponding percentage was $25\%$.

\vspace{-0.15cm}
\subsection{Movielens dataset}
\label{secNumExp}
\begin{figure}
        \begin{subfigure}[b]{.5\linewidth}
            \centering \includegraphics[scale=.175]{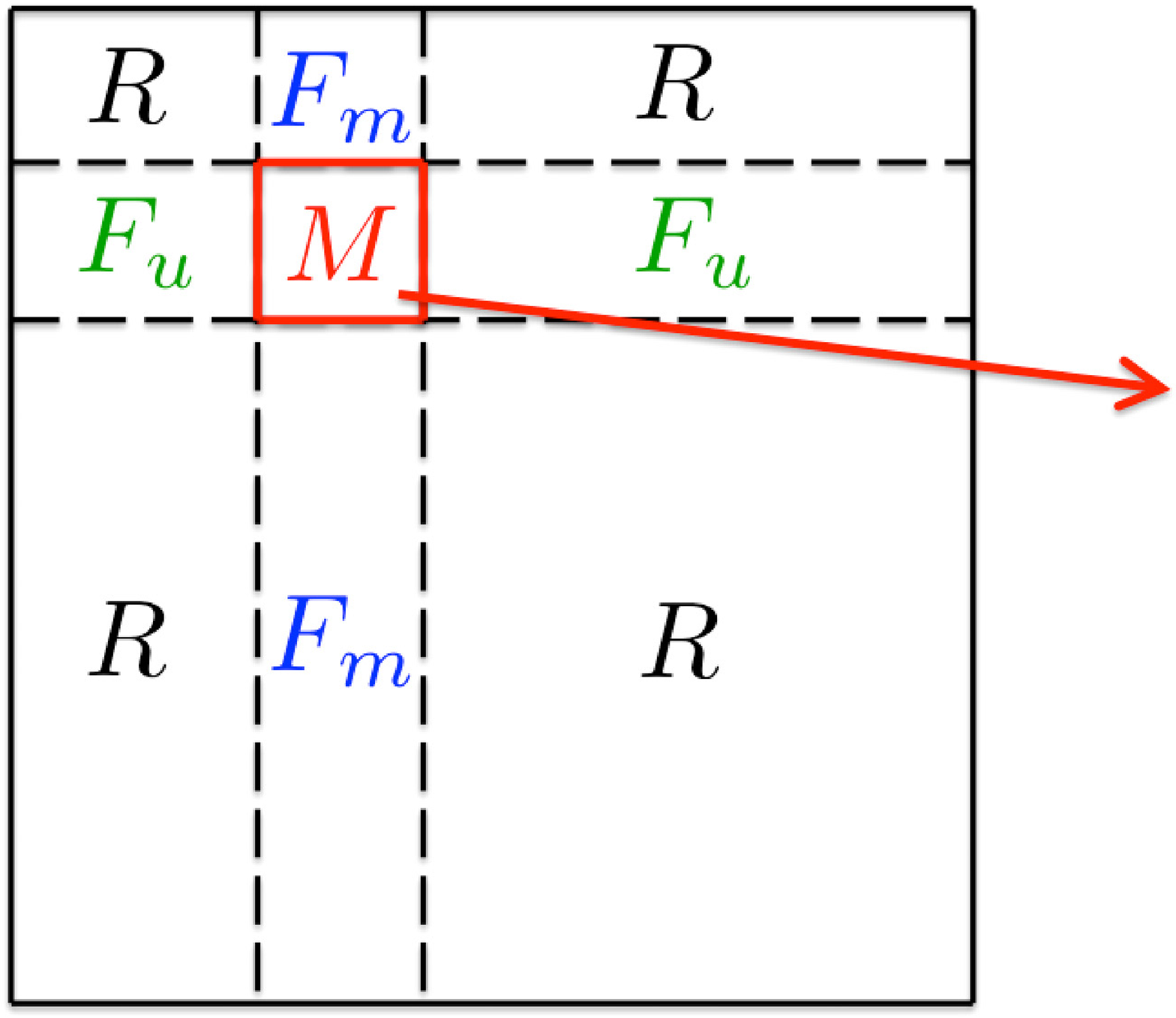}
	\vspace{-0.25cm}
            \vspace{0.65cm}
            \caption{Full Movielens matrix $A$.}\label{fig:movielens_blocks}
	\vspace{-0.15cm}
          \end{subfigure}
          \begin{subfigure}[b]{.5\linewidth}
            \centering\includegraphics[scale=.3]{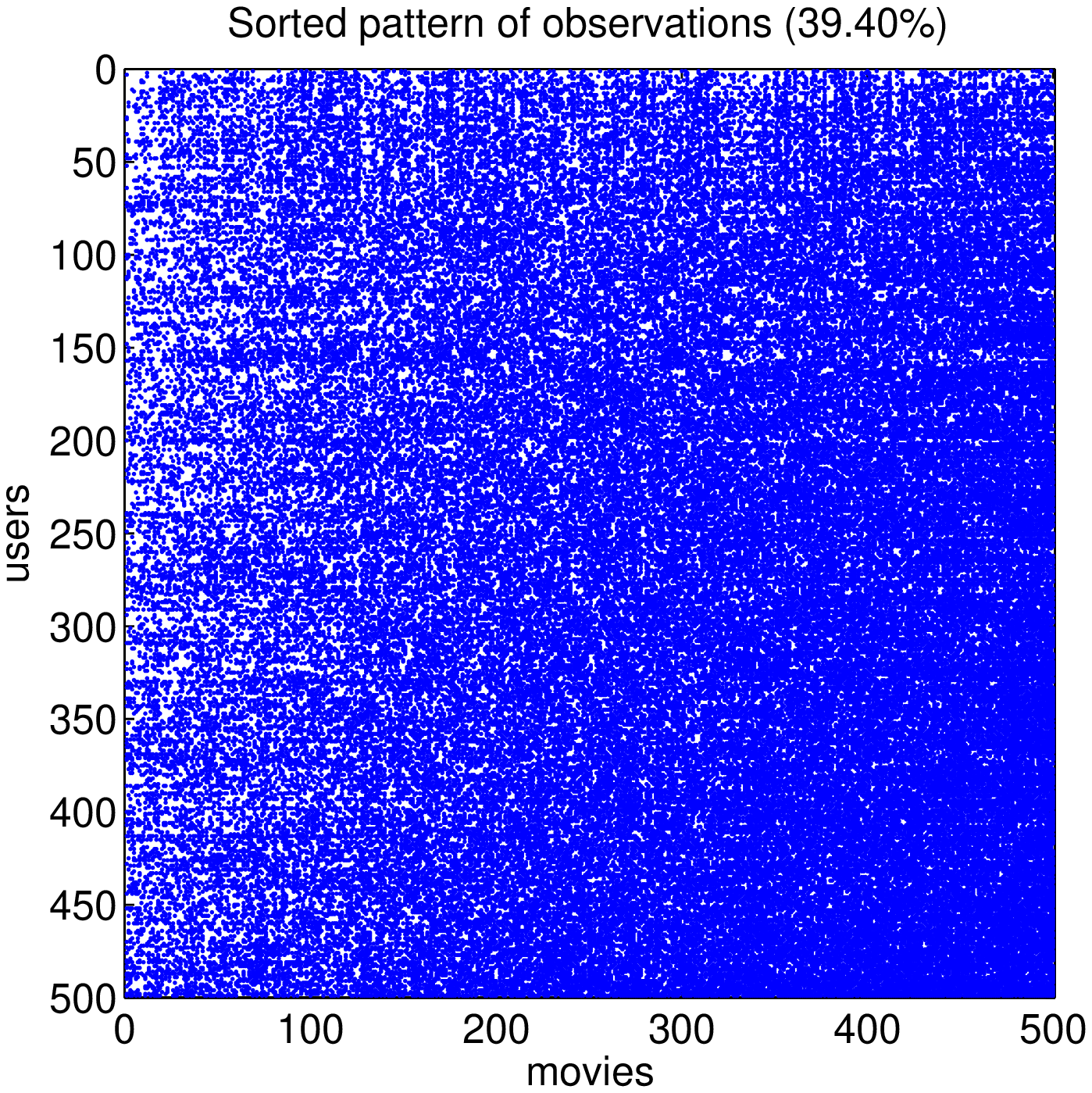}
	\vspace{-0.2cm}
            \caption{Submatrix $M\subset A$.}\label{fig:part_movielens}
	\vspace{-0.15cm}
          \end{subfigure}
          \caption{Movielens 10M dataset. The submatrix $M$ of $A$ is used for training and testing. The blocks $F_m$ and $F_u$ are used to construct the movie and user graphs.}\label{fig:movielens}
	\vspace{-0.25cm}
\end{figure}

\vspace{-0.25cm}
In this section, we report experiments on real data, which appear consistent with the results on the aforementioned artificial data. We work with the widely used \textit{MovieLens 10M} dataset \cite{pro:Miller03MovieLens}, containing ratings (`stars') from $0.5$ to $5.0$ (increments of $0.5$) given by 71,567 users for 10,677 movies. The density of the observations is $1.31\%$. In our experiments, we use a $500\times 500$ subset of the original matrix for the reconstruction evaluation. This serves two purposes: firstly, we can choose an arbitrary density of the submatrix, and secondly, we can use ratings outside of it as features for the construction of the column and row graphs, as detailed below (see Figure \ref{fig:movielens_blocks}). Furthermore, the effect of non-uniformity is weaker.\\
The density of the observations is selected as follows. We sort the rows (users) and columns (movies) by order of increasing sampling frequency (Figure \ref{fig:part_movielens}). Then, users and movies are chosen to be close to the $99$-th and $95$-th percentile of their corresponding distributions.\footnote{Since the number of movies in the full matrix is much smaller than the number of users, we keep more frequently rating users in order to have a dense features matrix when we create the users graph.} The resulting $500\times 500$ matrix has $39.4\%$ observed values  that correspond to the ratings that a user has given to a movie.
After a row and column permutation, the original MovieLens 10M matrix $A$ is partitioned in blocks $A = [M, F_u; F_m, R]$, where $M$ is the $500\times 500$ matrix that we use for our experiments (Figure \ref{fig:movielens_blocks}). We treat $F_u$ as the users feature matrix, $F_m$ as the movies feature matrix and discard the remaining matrix $R$.

\begin{figure}
        \begin{subfigure}[b]{.5\linewidth}
            \centering \includegraphics[scale=.38, trim=0 0 0 0, clip=true]{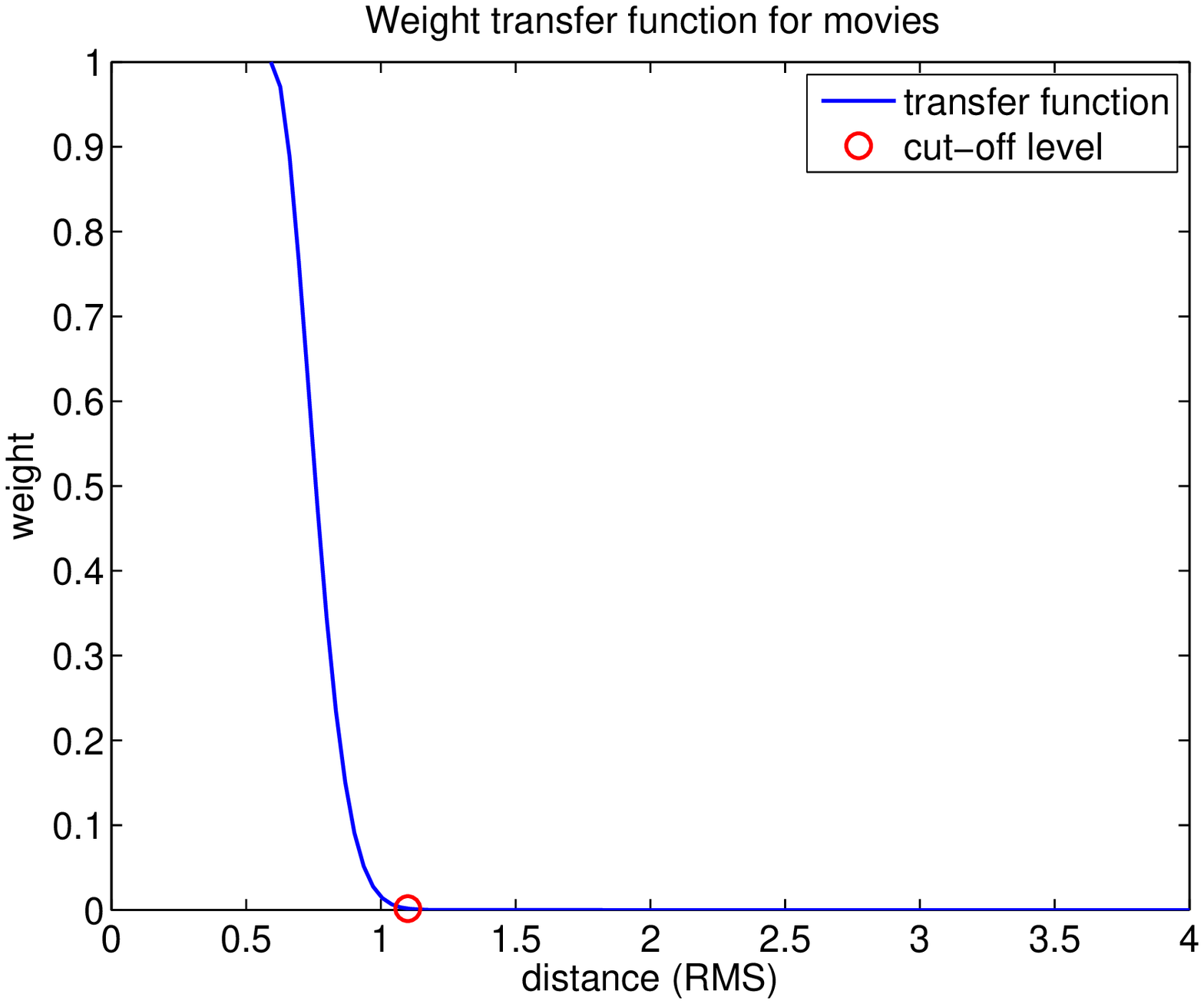}
            \caption{Transfer function used for movies graph.}\label{fig:transfer_fun_movies}
          \end{subfigure}
          \begin{subfigure}[b]{.5\linewidth}
            \centering\includegraphics[scale=.38, trim=40 0 0 0, clip=true]{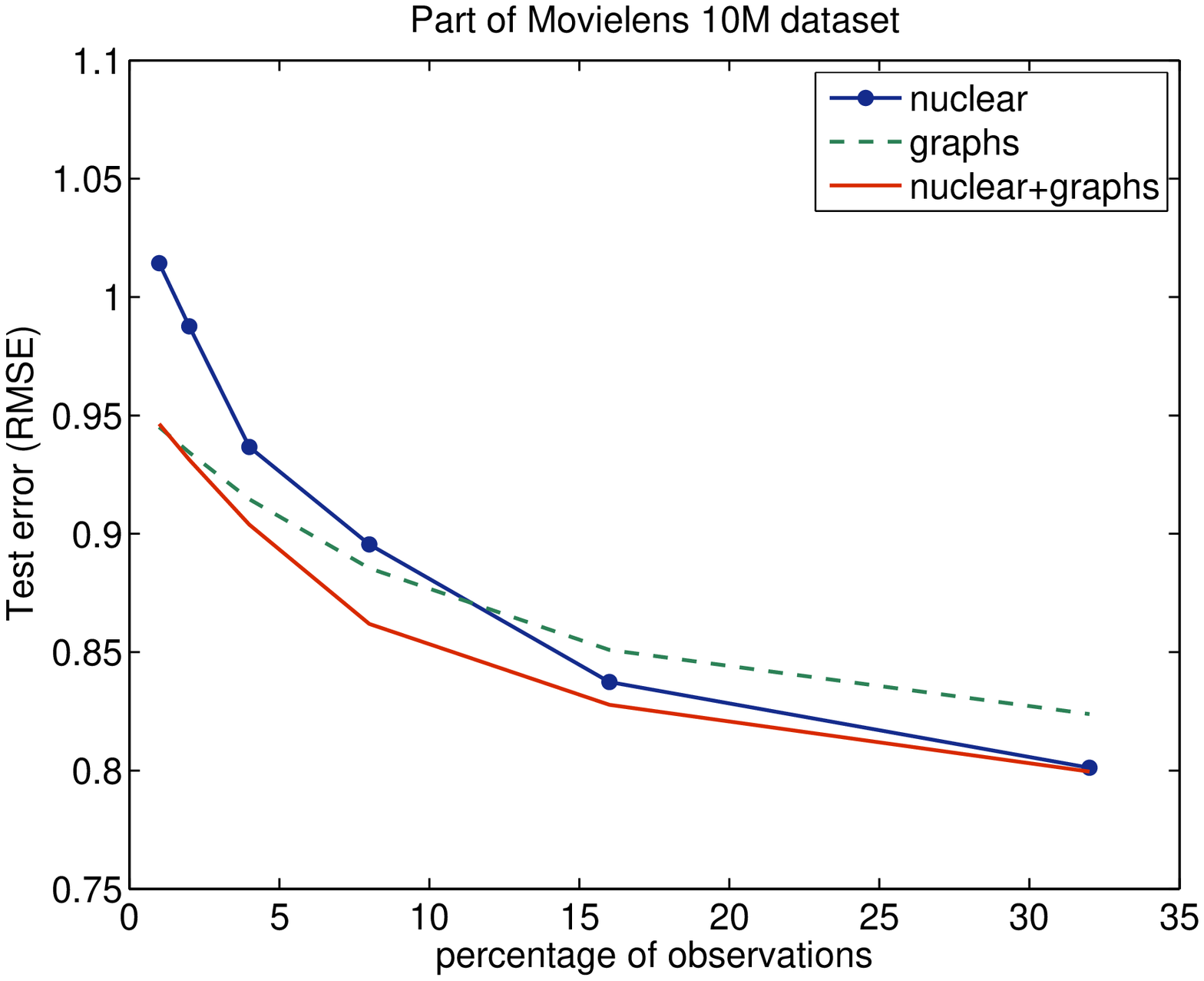}
            \caption{Reconstruction error}\label{fig:real_final}
          \end{subfigure}
         \caption{Experiments on a part of the Movielens-10M dataset}\label{fig:data_real}
	\vspace{-0.25cm}
\end{figure}

{\bf Graph construction.}
Quality of graphs obviously plays an important role to our matrix recovery algorithm. Since a detailed analysis of how to construct good graphs is beyond the scope of this paper, we will resort to a simple, yet natural way of constructing the graphs for our setting, using the feature matrices $F_u$ and $F_m$. We adapt the basic algorithm of \cite{belkin2003laplacian} to our setting that contains missing values. \\
The distance we use between two users is the RMS distance between their commonly rated movies
$d_{u_{ij}} = \|\left[F_{u_i} - F_{u_j}\right]_{\Omega_{u_{ij}}}\|_{\ell_2} / \sqrt{|\Omega_{u_{ij}}|}, \qquad\Omega_{u_{ij}} = \Omega_{u_i} \cap \Omega_{u_j},$ where $\Omega_{u_i}$ is the set of observed movie ratings for user (row) $i$ in $F_u$ and $|\Omega_{u_{ij}}|$ is the number of movies in $F_u$ that both users $i$ and $j$ have rated. We do the same to construct the movie distances from $F_m$, that is, for each pair of movies we only take into account the ratings from users that have rated both. Note that distances between movies or between users, that take values from $[0, 4.5]$ stars, share the same scale with the ratings and with the reconstruction error. Since the distances are all Euclidean, choosing the parameters of the graphs becomes more natural. The first choice we make is to use an $\epsilon$-neighborhood graph instead of a $k$-NN graph. To give weights to the edges, we use a Gaussian kernel, that is, $w_{u_{ij}} = \exp\left[-\left(d_{u_{ij}} - d_{\text{min-}u}\right)^2/\alpha\right] \text{ if } d_{u_{ij}} < \epsilon, 0 \text{ otherwise}$. In the latter, $d_{\text{min-}u}$ denotes the minimum distance among all pairs of users and $\alpha$ controls how fast the weights decay as distances increase. The transfer function used for the movies graph is plotted in Fig. \ref{fig:transfer_fun_movies}, while the one for users is nearly identical.\\
We give weight values equal to $1$ for distances close to the minimum one (around $0.6$ stars), while the weights decay fast as the distance increases. We choose $\epsilon = 1.1$ star, while $\alpha$ is chosen so that the transfer function is already very close to $0$ for $d_{u_{ij}}\rightarrow \epsilon$. This means that our model is equivalent to an $\text{inf}$-NN graph with the same exponential kernel. Note that the final reconstruction error is better than $1.1$ star in RMS, which justifies that distances that are smaller than that are trusted. We found that the results are indeed much better when a $k$-NN graph is not used. A possible explanation for this is that in that case a user that deviates a lot from the habits of other users would still have $k$ connections. These connections would not contribute positively in the recommendations quality regarding this user. \\ 
All this being said, it is here essential to emphasize that the graphs constructed for these experiments are not optimal. We foresee that the results presented in this paper can be further improved if one has access to detailed profile information about users and movies/products. This information is available to typical business companies that sell products to users.

{\bf Results. }
We apply a standard cross-validation technique to evaluate the quality of our completion algorithm. For this purpose, the $39.4\%$ observations of the $500\times 500$ matrix are split into a fixed test set ($7.4\%$) and a varying size training set (from $1\%$ to $32\%$). We perform $5$-fold cross validation to select the parameters $\gamma_n$, $\gamma_r$ and $\gamma_c$ of our model~\eqref{eq:problem2} and only use the test set to evaluate the performance of the final models.
The recommendation error results are plotted in Fig. \ref{fig:real_final}. 
The behavior of the algorithms is similar to the one exhibited by the noisy artificial data above (medium quality of graphs). For most observation levels our method combining nuclear norm and graph regularization (red line) clearly outperforms the rest. There are however two boundary phases that are noteworthy. When very few observations are available ($1\%$) there seems to be no benefit in adding the expensive nuclear norm term in the optimization problem, as the graph regularization alone (green line) performs best. On the other hand, for very dense observation levels ($32\%$) the nuclear norm (blue line) reaches the performance of the combined model. In general our combined model is very robust to observation sparsity, while the standard nuclear norm model performs worse even than the much cheaper graphs-only model for up to $8\%$ observations.

\vspace{-0.1cm}
\section{Conclusion}
\label{secCon}
\vspace{-0.15cm}
The main message of this work is that the standard low-rank matrix recovery problem can be further improved using similarity information about rows and columns. We solve an optimization problem seeking a low-rank solution that is structured by the proximity between rows and columns that form communities. As an application, our matrix completion model offers a new recommendation algorithm that combines the traditional collaborative filtering and content-based filtering tasks into one unified model. The associated convex non-smooth optimization problem is solved with a well-posed iterative ADMM scheme, which alternates between nuclear proximal operators and approximate solutions of linear systems. Artificial and real data experiments are conducted to study and validate the proposed matrix recovery model, suggesting that in real-life applications where the number of available matrix entries (ratings) is usually low and information about products and people taste is available, our model would outperform the standard matrix completion approaches. 
Specifically, our model is robust to graph construction and to non-uniformly sampling of observations. Furthermore, it significantly outperforms the standard matrix completion when the number of observations is small. \\
The proposed matrix recovery algorithm can be improved in several ways. The effect of the non-uniformity of sampling matrix entries, as discussed in Sections \ref{secMatComOnGrap} and \ref{secNumExp}, can be partially alleviated using a special weighting of the nuclear norm \cite{salakhutdinov2010collaborative}. The non-uniform sampling of user data points and movie data points from the corresponding manifolds, which influences the quality of graph Laplacians, can also be corrected using special graph normalizations \cite{art:CoifmanLafon06DifMap}. Furthermore, the optimization algorithm can be improved, firstly in terms of speed by using enhanced iterative schemes like \cite{nesterov2013first}. Secondly in terms of scalability, either by using distributed schemes like \cite{mardani2012distributed} or 
 by carrying out techniques from the recent work \cite{hsieh2014nuclear}, which deals with nuclear norm for matrices with sizes much bigger than the Netflix dataset. 


 \denserrlistbib
\bibliography{my_references}
\bibliographystyle{plain}



\end{document}

%% file: vassilis_macros.tex
\newcommand{\vect}[1]{\boldsymbol{#1}} 
\newcommand{\mat}[1]{\boldsymbol{#1}} 
\newcommand{\tvect}[1]{\tilde{\boldsymbol{#1}}} 
\newcommand{\tmat}[1]{\tilde{\boldsymbol{#1}}} 
\newcommand{\hvect}[1]{\hat{\boldsymbol{#1}}} 
\newcommand{\hmat}[1]{\hat{\boldsymbol{#1}}} 
\newcommand{\bvect}[1]{\bar{\boldsymbol{#1}}} 
\newcommand{\bmat}[1]{\bar{\boldsymbol{#1}}} 

\newcommand{\dummystring}{QWERTYU}
\newcommand{\vci}[3][\dummystr]{\ifthenelse{\equal{#1}{\dummystring}}{\vect{#2}_{#3}}{\vect{#2}_{#3}^{(#1)}}}
\newcommand{\mx}[3][\dummystr]{\ifthenelse{\equal{#1}{\dummystring}}{\mat{#2}_{#3}}{\mat{#2}_{#3}^{(#1)}}}
\newcommand{\tvci}[3][\dummystr]{\ifthenelse{\equal{#1}{\dummystring}}{\tvect{#2}_{#3}}{\tvect{#2}_{#3}^{(#1)}}}
\newcommand{\tmx}[3][\dummystr]{\ifthenelse{\equal{#1}{\dummystring}}{\tmat{#2}_{#3}}{\tmat{#2}_{#3}^{(#1)}}}
\newcommand{\hvci}[3][\dummystr]{\ifthenelse{\equal{#1}{\dummystring}}{\hvect{#2}_{#3}}{\hvect{#2}_{#3}^{(#1)}}}
\newcommand{\hmx}[3][\dummystr]{\ifthenelse{\equal{#1}{\dummystring}}{\hmat{#2}_{#3}}{\hmat{#2}_{#3}^{(#1)}}}
\newcommand{\bvci}[3][\dummystr]{\ifthenelse{\equal{#1}{\dummystring}}{\bvect{#2}_{#3}}{\bvect{#2}_{#3}^{(#1)}}}
\newcommand{\bmx}[3][\dummystr]{\ifthenelse{\equal{#1}{\dummystring}}{\bmat{#2}_{#3}}{\bmat{#2}_{#3}^{(#1)}}}

\DeclareMathOperator{\tr}{tr}

\DeclareMathOperator*{\argmin}{arg min}

\newcommand{\denserrlistbib}{
    \itemsep -1pt\topsep-6pt\partopsep-6pt
}